\def\year{2019}\relax
\documentclass[letterpaper]{article} 
\usepackage{aaai19}  
\usepackage{times}  
\usepackage{helvet} 
\usepackage{courier}  
\usepackage[hyphens]{url}  
\usepackage{graphicx} 
\usepackage{graphicx}  
\title{Visual Evaluation of Generative Adversarial Networks for Time Series Data}
\author{Hiba Arnout,\textsuperscript{\rm 1,2} Johannes Kehrer,\textsuperscript{\rm 1} Johanna Bronner,\textsuperscript{\rm 1} Thomas Runkler\textsuperscript{\rm 1,2}
\\ 
\textsuperscript{\rm 1}{Siemens AG, Corporate Technology, Otto-Hahn-Ring 6, 81739 Munich, Germany}\\
\textsuperscript{\rm 2}{Technical University of Munich, Department of Computer Science, Boltzmannstraße 3, 85748 Garching, Germany}\\
\{hiba.arnout, kehrer.johannes, johanna.bronner, thomas.runkler\}@siemens.com 
}
\begin{document}

\maketitle

\begin{abstract}
A crucial factor to trust Machine Learning (ML) algorithm decisions is a good representation of its application field by the training dataset. 
This is particularly true when parts of the training data have been artificially generated to overcome common training problems such as lack of data or imbalanced dataset. 
Over the last few years, 
Generative Adversarial Networks (GANs) have shown remarkable results in
generating realistic data. 
However, this ML approach lacks an objective function to evaluate the quality of the generated data. Numerous GAN applications focus on generating image data mostly because they can be easily evaluated by a human eye. 
Less efforts have been made to generate time series data. 
Assessing their quality is more complicated, particularly for technical data. 
In this paper, we propose a human-centered approach supporting a ML or domain expert to accomplish this task using Visual Analytics (VA) techniques. 
The presented approach consists of two views, namely a GAN Iteration View showing similarity metrics between real and generated data over the iterations of the generation process and a Detailed Comparative View equipped with different time series visualizations such as TimeHistograms, to compare the generated data at different iteration steps. 
Starting from the GAN Iteration View, the user can choose suitable iteration steps for detailed inspection. 
We evaluate our approach with a usage scenario that enabled an efficient comparison of two different GAN models.
\end{abstract}
\section{Introduction}
Time-dependent information arises in many fields ranging from meteorology, medicine to stock markets.
The analysis of such data is a central goal in VA, statistics, or ML and many related approaches exist \cite{analyze_ts,book_ts}.
In particular for ML methods a sufficient amount of training data and a balanced dataset (i.e. each data class is equally well represented) are crucial 
for a good performance. 
In reality, ML experts often face situations where these criteria are not satisfied.
In these cases, generating new data provides a possible solution.

This has pushed researchers to investigate new methods
for data generation. In this context, Generative Adversarial Networks
(GANs) \cite{GAN} are showing an outstanding performance. However, to trust a ML model e.g. a classifier trained on generated data, it is necessary to assess how realistic these data are and hence the performance of the generation process. 

Most efforts and best results have been shown for image generation, where the quality of the generated data can be easily assessed with the human eye.
Some research tried
to exploit this technique to generate time series data \cite{rgan}. 

GAN is a Minimax game between two
neural networks i.e. the Generator (G) and the Discriminator (D). D is a binary
classifier that tries to maximize its log-likelihood to distinguish between the real and the generated data. At the same time, G is trying to minimize the log-probability of the generated
samples that are recognized as false. 
if the data produced by the generator sufficiently represent the original data.
Much efforts are
made by researchers to discover suitable metrics to
evaluate the performance of GAN and can substitute a human judge.
The Discriminator
and Generator losses, for example, cannot be considered as a measure
of GAN performance 
and this ML approach lacks an objective function that defines an appropriate end of iteration with suitable 
data quality. 
Various evaluation methods have been described in \cite{Theis2016a} such as
Parzen window or Maximum Mean Discrepancy (MMD). As proven
in \cite{Theis2016a}, the use of these methods has various disadvantages.
Other methods e.g. inception score are designed only for images and
cannot be easily applied to time series data.
Therefore,
the quality of the generated data must be visually assessed by a
human judge \cite{rgan}.   
Finding common characteristics and differences
between real and generated data may be a complicated and
exhaustive task. In fact, comparing generated datasets to real
datasets is a challenge, especially for time series data. 

In this paper, we propose
a VA system to guide the ML expert in this evaluation
process for time series data. The developed framework presents a
method that make the real and generated data easily comparable
by combining VA (Colorfield, TimeHistograms) 
with algorithmic methods and enable the ML expert to trust the trained GAN model.
The contribution of the underlying work consists of 
an evaluation framework illustrated by two views namely, a GAN Iteration View and a Detailed Comparative View, that support ML and domain experts to assess the quality of time series data generated with GAN by providing:
\begin{itemize}

	\item An overview visualization that helps the analyst to identify interesting iteration of the GAN generation process.
	\item A comparison interface where the time series are visualized in a compact manner and ordered using Principal Components Analysis (PCA) to facilitate comparison by juxtaposition. 
\end{itemize}

 \section{Related Work}
GAN \cite{GAN} has gained a lot of attention in the last few years. Various
GAN models have been proposed such as DCGAN \cite{DCGAN}. Other
GAN models were designed for sequential data 
\cite{SeqGAN,mogren2016crnngan,rgan,MAD-GAN}.
Theis et al.\ \cite{Theis2016a} present an overview
of different evaluation methods for generative models with a
focus on images. These methods can be unreliable and cannot be easily applied to time series data.
According to Esteban et al.\ \cite{rgan} the evaluation of GAN is still an important problem. The authors present
two evaluation techniques for GANs generating time series: train on synthetic test on real
(TSTR) and train on real test on synthetic (TRTS) where the performance
of a model trained on synthetic generated data is evaluated
on real data and vice-versa. 
While their methods are mainly computational, our visual approach permits the user to explore the behavior of GAN over the iterations and to select the iteration with the best results. Also, we enable the user to check that the real data are equally well represented by the generated data which is a general goal for GAN.

Recently, different VA tools have been proposed for
Deep Learning (DL) models. Hohmann et al.\ \cite{sur_dl_va} give an overview about tools
developed for interpretability, explainability and debugging purposes
or to ease comparison between different DL models. 
A VA tool is introduced to understand GAN
\cite{gan_lab} equipped with two visualizations for D and G to give the user the opportunity to train GAN in the
browser and understand its functioning. 
An approach to understand GAN models and interpret them is presented by Wang et al.\ \cite{ganviz}. This method mainly focuses on explaining the behavior of GAN for domain experts. These approaches did not target the evaluation of the generated data and the considered GAN models were not designed to generate time series data.

Aigner et al.~\cite{analyze_ts} give an extensive overview of techniques for visualizing and analyzing
time series data. Numerous methods were proposed to perform predictive analysis on time series \cite{pred_va}. Van Wijk and Van Selow \cite{calendar}
describe an approach to recognize patterns or trends on different
time scales. The presented visualization illustrates the average of
each cluster of time series equipped with a calendar to highlight
the corresponding time scales. 
Muigg et al.\ \cite{kehrer} propose a focus+context method to interactively analyze a large number of function
graphs. 
 
TimeHistograms \cite{TimeHisto} are an extended version
of the standard histograms that allow a temporal investigation. We utilize this visualization in the Detailed Comparative View to highlight the time-dependent distribution of the generated data at different iterations and enable the user to check that GAN is capturing the distribution of the real data. 
Gogolou et al.\ \cite{comp_ts} compare three visualization
methods namely line charts, Horizon graphs and Colorfields in finding
and comparing similarities between time series. In this context,
multiple time series are compared to a unique time series identified
as a query. 
In our work, we employ Colorfields as a compact visualization for comparing many time series at different iterations using juxtaposition.
Javed et al.\ \cite{graph_perc} investigate the performance of small line graph,
braided graph, small multiples and horizon graphs in comparison,
slope and discrimination tasks. According to this study, shared
space techniques fit to a small number of time series, while split
space techniques are more suitable for a higher number of time series. 
Ferstl et al. \cite{Ferstl} analyze ensembles of iso-contours and use PCA to cluster similar ones. In contrast to their work, we use PCA to sort the time series in a consistent manner such that similar time series are located near each other.

\section{Design of Evaluation Framework}
In this section, we describe our 
evaluation framework, which supports ML experts in
generating time series data based on
a given number of real data. For the sake of simplicity, we consider
uni-variate time series of equal length. In this context, we present
a workflow that can be used to evaluate the performance
of GAN and assists the ML expert with the data generation. This
work is the result of an ongoing collaboration between ML
and VA experts. The feedback 
of practitioners that use GAN
to generate time series helped us in the framework's design. 
Our approach addresses some of the main issues with GAN training
that are frequently encountered by ML experts. The VA system fulfills the following design goals:
\begin{description}
	\item[Goal 1] Find iterations where an appropriate 
	behavior is achieved i.e. the iterations showing a sufficient quality of the generated
	data. Check if the number of iterations is sufficient or
	a higher number of iterations is needed.
	\item[Goal 2] 
	Compare the performance of different GAN models with different sets of parameters and support the ML expert in the decision making process to either trust or reject the current GAN model. Hence, the ML expert should be able to identify which GAN model and subsequently, which set of parameters is better.
	\item[Goal 3] Present an adequate method to visually evaluate the quality
	of the generated data i.e. detect if the data are noisy or show
	a different behavior compared to the real time series data. 
	ML experts should be able to decide wether the
	time series generated by a GAN algorithm are realistic.
	
	\item[Goal 4] Detect common GAN training problems such as non-convergence
	or mode collapse. Mode collapse is an important issue
	that a ML expert may encounter during training. In this case,
	the generator collapses to one mode and is not able to produce
	diverse samples. Almost all proposed GAN models \cite{GAN,DCGAN} suffer
	from this issue. 
	Our purpose is to offer the user the possibility to easily identify this phenomenon.
	Once the problem is detected, the ML expert can use existing techniques \cite{NIPS2016_6125} to improve the performance of the considered GAN model.
\end{description}
\begin{figure}
	\centering
	\includegraphics[scale= 0.29]{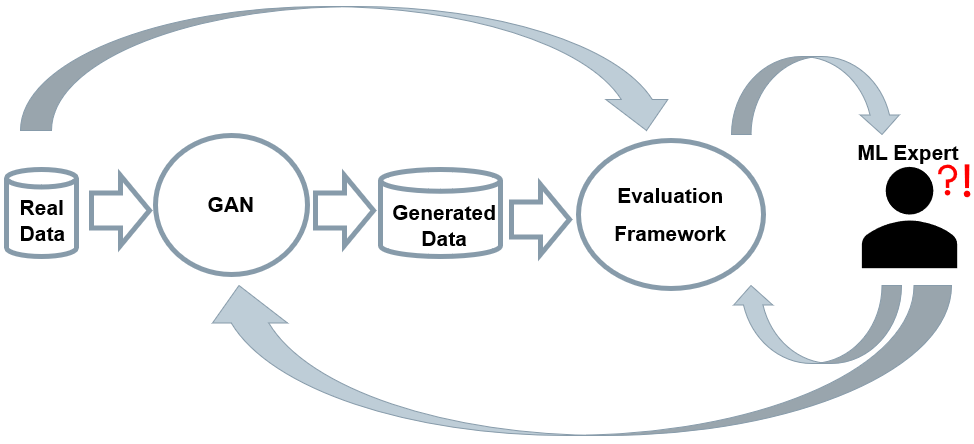}
	\caption{A visual evaluation workflow of GANs for time series data: The real and the generated data are integrated in the evaluation framework. The ML expert may interact with the evaluation framework to get more insight about the data and their properties. After a rigorous exploration of the data, he or she can decide to terminate the training process if the desired behavior is achieved. Otherwise, he or she has to run the GAN model with different parameters.}

	\label{fig:workflow}
\end{figure}
We design our framework based on these criteria. The ML expert
starts the process by generating time series with GAN. 
The evaluation framework is then used to check whether the
GAN model and the generated data fulfill the desired requirements.
If this is the case, the ML expert has succeeded to generate realistic data
and can stop the generation process. Otherwise he or she will have to
rerun the GAN model with different parameters and repeat the investigations in the framework. It should be noted that an online evaluation is also possible, i.e. the framework can be 
used during the training process. As the training process can take up to several days,
our approach may help to save valuable time by making sure during the training that
the GAN model is going in the right direction or restart the training
process if an unexpected behavior is detected. The proposed workflow is depicted in Fig.~\ref{fig:workflow}.

\section{Evaluation Framework Description}
Our proposed approach is characterized by two views: a GAN Iteration View
that gives the user a general impression about the behavior of GAN
over the iterations of the generation process and a Detailed Comparative View
 equipped with TimeHistograms \cite{TimeHisto}, Colorfields \cite{comp_ts} and line plots to further investigate particular time series selected by the user. The TimeHistogram displays the time-dependent distribution of all time series at a certain iteration. At the same time, the Colorfield
visualization allows further investigation and exploration of a multitude of generated time series at a certain iteration and compares them to the real time series. Moreover, a direct comparison between specific time series is made possible using the line plots visualization. 
To get more insights about the properties of the data, a measure of similarity and a dimensionality reduction technique
are used:
\begin{enumerate}
	\item Similarity measures such as Euclidean Distance (ED) or Dynamic Time Warping (DTW) are used as pairwise distance between two time series. 
	\item Principal Component Analysis (PCA) is used to arrange similar time series close to each other and facilitate comparison.
\end{enumerate}

\subsection{GAN Iteration View} The view consists of two components, namely the
incoming and outgoing nearest neighbor distances (see Fig.~\ref{fig:m1}a). The Incoming
Nearest Neighbor Distances (INND) represent for each generated time series the minimal distance to a real time series. The user can choose between ED and DTW as a distance measure. The
evolution of these minima throughout the iterations is in this
case shown. We repeat the same procedure calculating
the minima of each real time series to all generated time series.
This corresponds to the
Outgoing Nearest Neighbor Distances (ONND). 
A PCA is applied to the real time series data to transform the data points of each time series into uncorrelated components. 
The real data are then sorted based on the first principal component. 
To make both the real and generated data comparable, the same transformation is applied to the generated data.
A heatmap visualization is used to depict the
incoming and outgoing minimal distances. The intensity of the color
of each pixel highlights the value of the minimal distance.
A dark pixel represents a high distance value, while a
brighter pixel denotes a lower distance value. The
nearest neighbor distances give an overview about the overall performance
of GAN over the iterations and allow for different types of investigations:
\begin{itemize}
	\item Are the time series becoming more realistic with the iterations i.e. do the INND/ ONND become smaller?
	\item Are INND / ONND reaching a stable behavior and indicating nearly constant values?
	\item Is the variation in the real data representative for the generated data i.e. are all types of generated time series equally similar to the real data (INND) and are all real time series equally well represented by the different generated time series or do generated time series correspond to a limited number of real ones (ONND)?
\end{itemize}
The user can
interactively select interesting iterations in the GAN Iteration View and get more
insights about the selected iterations in the other view. This will
permit him to identify the iteration with the best behavior. For sake of simplicity, we use ED as a similarity measure in the rest of the paper.

\subsection{Detailed Comparative View} This view is equipped with
TimeHistograms \cite{TimeHisto} depicting the distribution of the real and the
generated data, a Colorfield visualization as a compact representation of the corresponding
time series and a Selected Samples View allowing comparison by superposition 
(see Fig.~\ref{fig:m1}b). In the Colorfield and TimeHistogram Views, the real data are shown on the left and
selected iterations of the generated data are shown on the right. This setup enables a comparison by juxtaposition between the real and the generated data as well as between different iterations of the generation process. 
The user can investigate different iterations at the
same time. Both real and generated data are automatically sorted for each iteration step based on the first principal component. 
The TimeHistograms enable a time-dependent
investigation of the distribution of the data and a comparison between
the distribution of the real and the generated data. 
This visualization represents a possibility to check if the model is working 
properly and capturing the distribution of the real data. The Colorfield
visualization is used to depict the time series and enables a rigorous exploration of the generated time series and
their properties. Each heatmap represents all the data of a specific iteration where each row corresponds to a time series. 
This visualization permits the
user to compare a high number of time series in an efficient manner. 
Additionally, a rigorous investigation of some selected time series is made possible with the Selected Samples View equipped with two visualizations. 
To give the ML expert more insights about the real data, the first plot depicts their median $med(r)$ and the amount of data falling in the $68 th$, $95 th$ and $99.7 th$ percentile denoted with $68 prct$, $95 prct$ and $99 prct$ respectively.
The user may add interesting, real or generated, time series to the plot to investigate their properties and compare them by superposition. Each generated and real time series is denoted with $g\_iter\_id$ and $r\_id$ respectively where $iter$ is the number of the iteration at which the time series was generated and $id$ is the index of the time series sorted with PCA. 
The second plot highlights the absolute value of the element-wise difference between the selected time series $r/g\_i$ and the median of the set of real data $med(r)$. 
This feature provides additional information about the selected data by directly comparing their behavior to a reference value, namely the median of the real data. 
Our main
concern is to enable an exploration of the behavior of the ML model
over the iterations and an investigation of the similarity between the
real and the generated data. Hence, the presented human-centered approach gives the opportunity to build a relationship of trust between the ML expert and the AI algorithm.  

\section{Use case}
\begin{figure*}
	\centering
	\includegraphics[scale= 0.41]{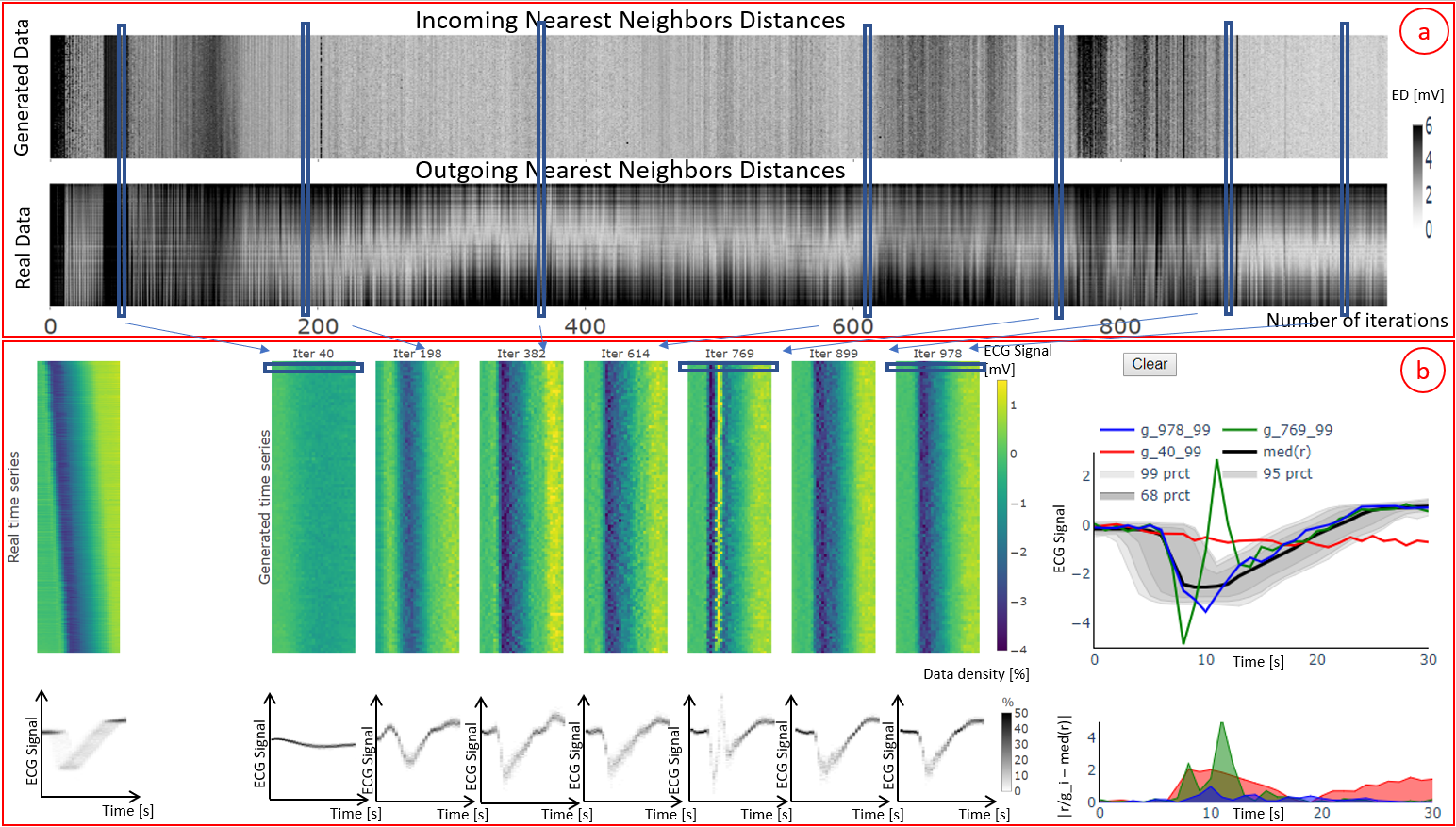}
	\caption{Results of a first GAN model generating time series. 
		The computed incoming and outgoing minimal distances are integrated in the GAN
		Iteration View (a). Selected columns in the GAN Iteration View, denoted with blue rectangles, are depicted in the Detailed Comparative View (b).}
	\label{fig:m1}

\end{figure*}
\begin{figure*}
	\centering
	\includegraphics[scale= 0.41]{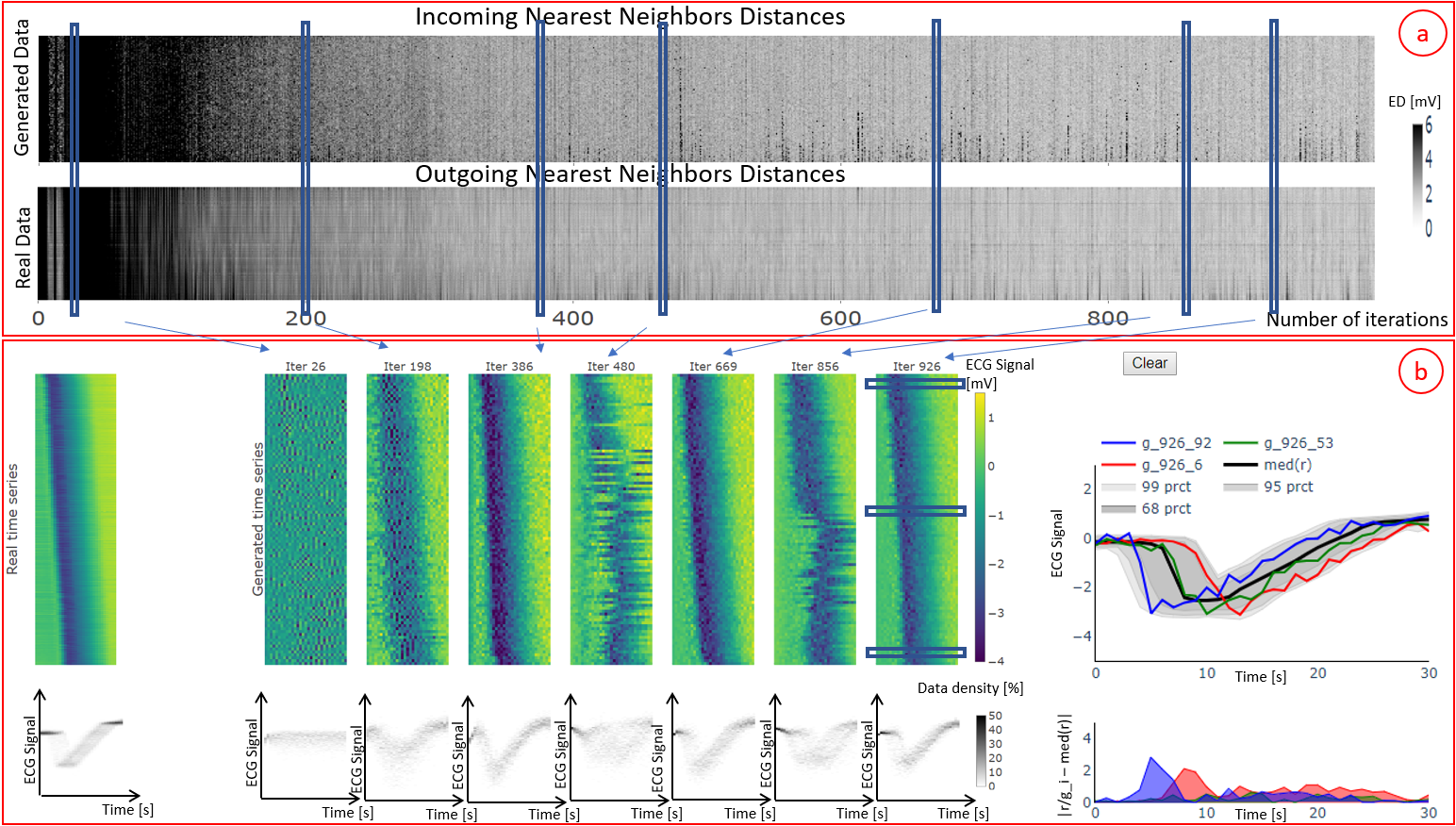}
	\caption{Results of a second GAN model obtained by tuning the parameters of the first model. In comparison to model 1, this model is showing a more stable and smooth behavior in terms of the incoming and outgoing nearest neighbor distances. The Colorfields, depicted in the Detailed Comparative View (b), indicate that the last iteration is reproducing the shift present in the real data and its TimeHistogram is similar to the TimeHistogram of the real data.}
	\label{fig:m2}

\end{figure*}
To demonstrate the utility of the developed framework, our ML expert
tested the proposed method on a GAN model \cite{mogren2016crnngan} to generate data
based on the real dataset \cite{PhysioNet}. The considered dataset consists of 7
long-term Electrocardiogram (ECG) for a period of 14 to 22 hours
each. It contains two classes depicting the normal and
abnormal behavior. To reduce the training time, only 30 time points from the real time series are considered.
The ML expert used one class in his experiments.
In our case, the performance of GAN is evaluated for two different
parameter configurations, namely model 1 and model 2. 
The corresponding results
are depicted in Fig.~\ref{fig:m1} and \ref{fig:m2}. 
\begin{figure}[h]
	\centering
	\includegraphics[scale= 0.7]{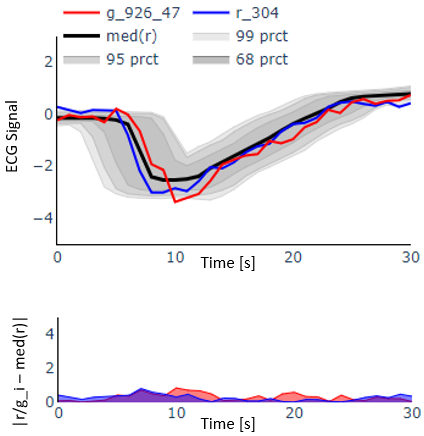}
	\caption{Illustration of the Selected Samples View with the median of the real data $med(r)$, $68th$, $95th$ and $99th$ percentile denoted with $68 prct$, $95 prct$ and $99 prct$ respectively, time series $g\_926\_47$ generated at iteration 926 by model 2 and a real time series $r\_304$. The absolute value of the element-wise differences of $g\_926\_47$ and $r\_304$ to the median $med(r)$ are denoted in red and blue respectively. The time series $g\_926\_47$ is falling in the $98th$ percentile of the real data and $g\_926\_47$ and $r\_304$ are showing a similar behavior.}
	\label{fig:pl_ts}
	
\end{figure}

The GAN Iteration View (Figs.~\ref{fig:m1}a and~\ref{fig:m2}a) depicts the variation of
INND and ONND depending on the iterations.
The first iterations are characterized by high INND and ONND. As the number of iterations increases,
an improvement in terms of INND can be
seen. Hence, the generated data are progressively reaching similar
values to the original data and the performance of the ML algorithm
is increasing with a growing number of iterations. However, the
ED values corresponding to some iterations
in model 1 sharply increase. Model 2 is showing a more stable
behavior. In fact, after approximately 300 iterations, the INND are almost constant. ONND in Fig.~\ref{fig:m1}a show that the
values of the ED at the top and bottom of the view are
still high. As the real time series are sorted with PCA, our expert concludes
that the real time series with an important shift are characterized
by a high outgoing minimal distance. Hence, the time
series produced by the first model are similar to a specific type of
the real time series namely the time series that are in the middle. He
hypothesizes that this GAN model was not able to reproduce the shift
present in the real data and is collapsing to one mode. In contrast to
model 1, ONND illustrated in Fig.~\ref{fig:m2}a depicts a low outgoing ED
for all real data. ONND helped the ML
expert to verify that the generated data are diverse and do not correspond
to a specific type of time series but to almost all real ones. 

Afterwards, the ML expert selects some interesting columns in the
GAN Iteration View and continues his investigation in the other view. 
For both scenarios,
the user selected an iteration at the begin of the training process,
certain columns with low EDs in the middle, few
columns characterized by high INND and ONND in model 1 and 2 and some columns showing a stable behavior
within the last hundred iterations. Initially, the time-dependent
distribution of the generated data was completely different from the
real data and noise was generated. An improvement in the performance
is noticeable after approximately 200 iterations. In general,
the time-dependent distribution and the quality of the generated data
are becoming more realistic over the iterations. An enhancement in
the results is observed between the iterations 382, 614 and 899 for
model 1 and the iterations 386 and 669 for model 2. 
To inspect the behavior of model 1 rigorously, the user selected some time series generated at different iterations.
In the Selected Samples View, he noticed that at iteration 764 the generated data presents a strange peak and at iteration 40 noise is generated. 
Hence, the evaluation framework helped the ML expert to detect if the data are noisy or have a
different behavior from the real data (Goal 3).

To conclude, GAN
was not able to generate realistic time series in the first iterations at all and is learning
the properties and features of the real time series over the time.
However, the data quality can decrease drastically after one iteration i.e.
iterations 769 and 480 in the first and second scenario respectively.
The ML expert confirms that this is an expected behavior with neural
networks because their performance is not monotonic. An inspection of the last hundred iterations allows the ML expert
to find an iteration with the best result (Goal 1). This corresponds
to iteration 978 for model 1 and iteration 926 for model 2. In both
cases, the generated data are smooth and realistic. 
However, the TimeHistogram of the data generated with model 1 is still different
from the TimeHistogram of the real data. Moreover, the Colorfield
View demonstrates that the samples are not as diverse as in the real
data.  A rigorous investigation of these time series in the Selected Samples View shows that all the generated data are falling in the 68th percentile of the real data and are too close to the median i.e. their difference to the median of the real data is low.
This confirms the hypothesis of the ML expert when he
observed the GAN Iteration View. Thus, the user was able to easily detect the
mode collapse phenomenon, one of the hardest training problems for
GAN (Goal 4). 
In order to avoid this problem, the ML expert 
used in model 2 a
normal distributed noise instead of the uniformly distributed noise
and applied a technique introduced in \cite{NIPS2016_6125} namely mini-batch discrimination.
In contrast to model 1, we clearly see that model 2 is reproducing the distribution of the real data much better. 
For further exploration, the ML expert selected different time series from the Colorfield View and visualize them in the Selected Samples View.
He noticed that model 2 is reproducing the shift present in the real data. This model is generating time series that are moved to the right and to the left and are characterized, at some data points, with a high difference to the median. As a last step, our expert used the Selected Samples View to directly compare the generated and real data. Fig.~\ref{fig:pl_ts} shows a real and a generated time series selected by the ML expert.
We clearly see that the behavior of the generated time series is similar to the behavior of the real data.
Hence, the second GAN model presents a more realistic behavior and was able at iteration 926 to generate time series that are rare in the real dataset. 
The ML expert concludes that model 2 is achieving the desired behavior. Hence,
the proposed framework helped the ML experts to find
a trustworthy GAN model with a set of parameters producing the best results (Goal 2).

Finally, the ML
expert said that it was helpful to see the evolution of the behavior
of GAN over the iterations and how the similarity between the real
and the generated data is improved with the number of iterations.
He was able to assess the
quality of the generated data and find a reliable GAN model achieving trustworthy results.

\section{Conclusion}
In this work, we proposed a visual approach to evaluate and optimize GAN models generating time series data. The proposed method
is based on two visualization techniques namely Colorfield
and TimeHistogram as well as a distance measure. 
The distance measure is used in a sophisticated manner
to compute the incoming and outgoing nearest neighbor distances. The VA system supports ML experts in the evaluation process.
The utility of the developed framework is demonstrated
with a real-world use case where the ED is used as a distance measure. In this case, a ML expert evaluated the performance of two different GAN models in generating time series
based on existing real ones.
He was able to detect that the first GAN model 
generates samples which are not diverse. 
This corresponds to one of two criteria 
which are verified by the inception score \cite{noteincep}, a metric to assess the quality of images generated with GAN.
The second criteria states that the
mapping between the generated and the real data must be
clear, i.e. the generated samples should correspond to an easily recognizable class. 
This aspect will be further considered in a forthcoming publication. 
Other developments are planned to allow for increased transparency and deeper understanding of the GAN algorithm such as: additional views that highlight the decision making process of the discriminator and an efficient comparison between data generated from different GAN models. 
Moreover, we plan to consider other GAN architectures and other real-world use cases that will involve the feedback of different ML and domain experts in a forthcoming publication. The presented work constitutes a starting point to guide a human to decide if data generated by a GAN algorithm can be used to build reliable and trustworthy Artificial Intelligence (AI) models. 
We believe that this topic will gain in importance in the future since more AI algorithms will rely on generated data.


%

%
\bibliography{template}
\bibliographystyle{aaai}
\end{document}